\documentclass[11pt]{article}

\usepackage{microtype}
\usepackage{booktabs}
\usepackage{url}

\usepackage{amsmath}
\usepackage{amsthm}
\usepackage{graphicx}
\usepackage{longtable}
\usepackage{multirow}
\usepackage{subcaption}
\usepackage{listings}

\usepackage{xcolor}

\definecolor{codegreen}{rgb}{0,0.6,0}
\definecolor{codegray}{rgb}{0.5,0.5,0.5}
\definecolor{codepurple}{rgb}{0.58,0,0.82}
\definecolor{backcolour}{rgb}{1,1,1}

\lstdefinestyle{mystyle}{
    backgroundcolor=\color{backcolour},   
    commentstyle=\color{codegreen},
    keywordstyle=\color{magenta},
    numberstyle=\tiny\color{codegray},
    stringstyle=\color{codegreen},
    basicstyle=\ttfamily\footnotesize,
    breakatwhitespace=true,         
    breaklines=true,                                 
    tabsize=2
}
\newcommand\myparagraph[1]{\vspace{6pt}\noindent\textbf{#1}\quad}

\usepackage[final,workshop]{automl}

\usepackage{automl}
\usepackage{booktabs}

\usepackage{natbib}
\title{PMLBmini: A Tabular Classification Benchmark Suite for Data-Scarce Applications}

\author[1]{\nameemail{Ricardo Knauer}{ricardo.knauer@htw-berlin.de}}
\author[1]{\nameemail{Marvin Grimm}{marvin.grimm@htw-berlin.de}}
\author[1]{\nameemail{Erik Rodner}{erik.rodner@htw-berlin.de}}

\affil[1]{KI-Werkstatt, University of Applied Sciences Berlin, Germany}

\hypersetup{%
  pdfauthor={Ricardo Knauer, Marvin Grimm, Erik Rodner},
  pdftitle={PMLBmini: A Tabular Classification Benchmark Suite for Data-Scarce Applications},
  pdfsubject={AutoML},
  pdfkeywords={benchmark, data scarcity, AutoML, deep learning, meta-learning}
}

\begin{document}

\maketitle

\begin{abstract}
In practice, we are often faced with small-sized tabular data. However, current tabular benchmarks are not geared towards data-scarce applications, making it very difficult to derive meaningful conclusions from empirical comparisons. We introduce PMLBmini, a tabular benchmark suite of 44 binary classification datasets with sample sizes $\leq$ 500. We use our suite to thoroughly evaluate current automated machine learning (AutoML) frameworks, off-the-shelf tabular deep neural networks, as well as classical linear models in the low-data regime. Our analysis reveals that state-of-the-art AutoML and deep learning approaches often fail to appreciably outperform even a simple logistic regression baseline, but we also identify scenarios where AutoML and deep learning methods are indeed reasonable to apply. Our benchmark suite, available on \url{https://github.com/RicardoKnauer/TabMini}, allows researchers and practitioners to analyze their own methods and challenge their data efficiency.
\end{abstract}

\section{Introduction} \label{sec:intro}

The easy access to data has fueled machine learning research in recent years. Massive text corpora crawled from the web have given rise to large language models such as GPT-3 with emergent abilities such as in-context learning \citep{brown2020language}. Large-scale image data have served as the foundation for text-to-image systems like DALL-E 3 \citep{betker2023improving}, large-scale video data for text-to-video generators like Sora \citep{videoworldsimulators2024}. In contrast to text, image, or video data, collecting large- or even medium-sized tabular data is often challenging in practice, despite tabular data being among the most ubiquitous dataset types in real-world applications \citep{borisov2022deep,mcelfresh2023neural,shwartz2022tabular}. GitTables, for example, a curated corpus of 1 million tables from GitHub, only contains 142 instances on average \citep{hulsebos2023gittables}. In clinical diagnostic or prognostic settings, the number of instances is frequently limited due to the rareness of medical conditions or patient losses at follow-up assessments, respectively \citep{moons2015transparent,moons2019probast,steyerberg2019clinical}.

The difficulty to acquire large- or even medium-scale tabular datasets in many domains presents researchers and practitioners with a unique set of challenges. On the one hand, overfitting is a major concern when applying complex algorithms on small-sized datasets. On the other hand, cross-validation folds may become too small to adequately represent both the original sample and the population of interest. This makes it very difficult to find good hyperparameter settings so that data-driven hyperparameter optimization may fail to increase, or may even decrease, the predictive performance for individual datasets in the low-data regime \citep{riley2021penalization,vsinkovec2021tune,van2020regression}. To facilitate empirical comparisons across studies in this setting, it is therefore imperative to systematically evaluate machine learning pipelines not on a hand-picked, narrow selection of datasets, but on a standardized, diverse dataset collection - a benchmark suite \citep{bischl2021openml,fischer2023openml,gijsbers2019open,gijsbers2022amlb,mcelfresh2023neural,olson2017pmlb,romano2022pmlb}.

\begin{figure}[t]
\centering
\includegraphics[width=\textwidth]{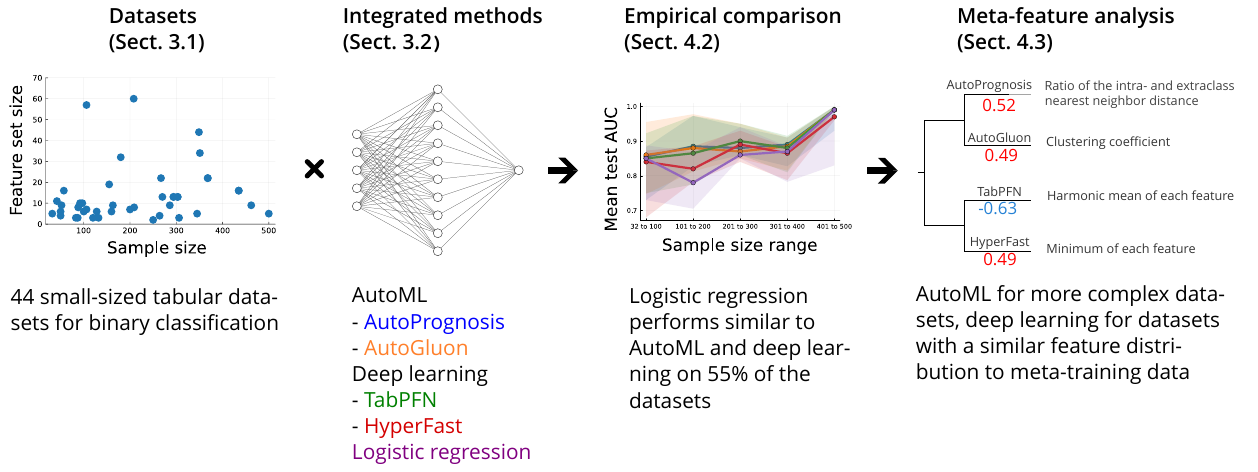}
\caption{Overview of our work on PMLBmini, the first tabular classification benchmark suite specifically for data-scarce applications.}
\label{fig:overview}
\end{figure}

In this paper, we contribute to the tabular benchmarking literature in the following ways:
\begin{enumerate}
\item We conduct a \textbf{narrative review on tabular benchmark suites for data-scarce applications} with sample sizes $\leq$ 500, and find that (very-)small-sized datasets are generally underrepresented in current tabular benchmarks (Sect.~\ref{sec:relatedwork}).

\item We introduce \textbf{PMLBmini, the first tabular benchmark suite specifically for the low-data regime with 44 binary classification datasets (Sect.~\ref{sec:datasets}), and use our suite to compare state-of-the-art machine learning methods, i.e., automated machine learning (AutoML) frameworks and off-the-shelf deep neural networks, against logistic regression (Sect.~\ref{sec:methods})}. Overall, we show that logistic regression performs similar to AutoML and deep learning approaches in terms of discrimination on 55\% of the datasets, and present the best L2-regularization hyperparameter obtained on each dataset that can be used for meta-learning in data-scarce applications (Sect.~\ref{sec:results}). We also conduct an extensive meta-feature analysis to assess under which conditions, i.e., dataset properties, AutoML and deep learning methods outperform a logistic regression baseline (Sect.~\ref{sec:metafeature}). Please refer to Fig.~\ref{fig:overview} for an overview of our work.

\item We \textbf{release PMLBmini for researchers and practitioners to benchmark their own tabular classifier, and to analyze if and when it succeeds or fails}. This way, we aim to equip the community with an easily accessible, practical set of tools for empirical evaluations in the low-data regime (Sect.~\ref{sec:interface}). 
\end{enumerate}

\section{Related Work and Desiderata} \label{sec:relatedwork}

Machine learning repositories for tabular data abound, prominent examples being Kaggle, UCI, or OpenML \citep{vanschoren2014openml}, which form the basis for a wide range of carefully curated tabular benchmark suites \citep{bischl2021openml,fischer2023openml,gijsbers2019open,gijsbers2022amlb,mcelfresh2023neural,olson2017pmlb,romano2022pmlb}. In spite of the practical relevance and unique challenges in the low-data regime (Sect.~\ref{sec:intro}), small-sized datasets with sample sizes $\leq$ 500 have received very little attention in benchmarking studies so far, though \citep{fischer2023openml,gijsbers2019open}. The suites that do include $\geq$ 1 small-sized tabular benchmark dataset are the Penn Machine Learning Benchmarks (PMLB) \citep{olson2017pmlb,romano2022pmlb}, the AutoML Benchmark (AMLB) \citep{gijsbers2022amlb}, the OpenML-CC18 \citep{bischl2021openml}, and TabZilla \citep{mcelfresh2023neural}. Their respective dataset sources, task types, sample size range, number of included datasets, and number of included datasets with sample sizes $\leq$ 500 are shown in Table~\ref{table:overview}. AMLB, OpenML-CC18, and TabZilla include an insufficient number of small-sized datasets to allow for meaningful comparisons in the low-data regime. PMLB, on the other hand, provides no interface for users to easily evaluate their own machine learning pipeline against a range of simple and state-of-the-art baselines on a preselected collection of small-sized datasets. We therefore formulate our desiderata for a tabular classification benchmark suite for data-scarce applications as follows:
\begin{itemize}
\item \textbf{Dataset size}: The collection should include only small-sized datasets with sample sizes $\leq$ 500, unlike any other tabular benchmark suite \citep{bischl2021openml,fischer2023openml,gijsbers2019open,gijsbers2022amlb,mcelfresh2023neural,olson2017pmlb,romano2022pmlb}. A cut-off at a sample size of 500 allows us to assess machine learning methods where the OpenML-CC18 leaves off \citep{bischl2021openml} and extend prior evaluations on this benchmark to the low-data regime \citep{bonet2023hyperfast,hollmann2023tabpfn,muller2023mothernet}.

\item \textbf{Dataset complexity}: To allow for assessing if and when state-of-the-art machine learning methods perform better than simple baselines, the suite should include both more and less complex datasets, and not exclude easy problems a priori \citep{bischl2021openml,fischer2023openml,gijsbers2019open,gijsbers2022amlb,mcelfresh2023neural}. In Sect.~\ref{sec:results}, we show that the inclusion of less difficult datasets does not prevent us from finding statistically significant performance differences between approaches when using our entire dataset collection.

\item \textbf{Collection size}: The suite should be sufficiently large to allow for thorough empirical comparisons. Small-sized datasets are typically underrepresented in current tabular benchmarks \citep{bischl2021openml,gijsbers2022amlb,mcelfresh2023neural} or not included at all \citep{fischer2023openml,gijsbers2019open}.

\end{itemize}

\tabcolsep=0.25cm
\begin{table}
  \centering
  \caption{Benchmark suites that include small-sized tabular datasets}
  \begin{tabular}{p{3.3cm} p{2cm} p{2cm} p{2cm} p{1.7cm} p{1.2cm}}
   \toprule 
    \textbf{Benchmark} & \textbf{Dataset sources} & \textbf{Task types} & \textbf{Sample size range} & \textbf{\# Datasets} & \textbf{\# Small-sized datasets}\\
   \midrule 
    PMLB \citep{olson2017pmlb,romano2022pmlb} & Kaggle, UCI, OpenML, and others & Classification, regression & 32 to over 1 million & 419 & 158 \\[1.1cm]
    AMLB \citep{gijsbers2022amlb} & OpenML & Classification, regression & 100 to 10 million & 104 & 2 \\[0.7cm]
    OpenML-CC18 \citep{bischl2021openml} & OpenML & Classification & 500 to 96320 & 72 & 1 \\[0.7cm]
    TabZilla \citep{mcelfresh2023neural} & OpenML & Classification & 148 to over 1 million & 36 & 4 \\[0.7cm]
    \midrule
    \textbf{PMLBmini (ours)} & OpenML & Classification & 32 to 500 & 44 & 44 \\
   \bottomrule
  \end{tabular}
  \label{table:overview}
\end{table}

\noindent As an additional preference, the suite should provide a user-friendly Python interface for researchers and practitioners to benchmark their own tabular classifier against baseline methods on the preselected dataset collection. Users should also be able to perform a comprehensive meta-feature analysis to find out under which conditions, i.e., dataset properties, their approach is better suited, which is not supported by most frameworks out of the box \citep{bischl2021openml,fischer2023openml,gijsbers2019open,olson2017pmlb,romano2022pmlb}. In the next section, we describe how we constructed our tabular classification benchmark suite for data-scarce applications based on our desiderata and preference.

\section{Benchmark Design}

In the following, we report on the design of our tabular classification benchmark, PMLBmini, starting with the included datasets. We then outline the baseline methods that we integrated into our suite, and finally provide details on how our benchmarking tool can be used for both in-depth empirical comparisons and meta-feature analyses in the low-data regime.

\subsection{Datasets} \label{sec:datasets}

We selected all binary classification datasets with sample sizes $\leq$ 500 from the curated benchmark suite with the largest number of small-sized datasets, PMLB (Sect.~\ref{sec:relatedwork}), resulting in 44 datasets (Table~\ref{table:results} in Appendix~\ref{sec:appendix_results}). There was no sensitive personally identifiable information or offensive content in the selected datasets, there were no missing values, and categorical features were already numerically encoded in PMLB. This provided a simple, extensible basis for our tabular classification benchmark. We shuffled all datasets and encoded all labels to \{0, 1\}. The 2 binary classification datasets from the TabZilla benchmark (\textit{colic} and \textit{heart-h}) are already included in PMLB and therefore in our suite, the 2 binary classification datasets from AMLB and OpenML-CC18 (\textit{arcene} and \textit{dresses-sales}, respectively) are not. Both \textit{arcene} and \textit{dresses-sales} would be clear outliers in our collection, though; the former because its feature set size is roughly 6000\% larger than the maximal feature set size in our suite, the latter because > 80\% of its instances contain missing values. Even without the 2 excluded datasets from AMLB and OpenML-CC18, our final tabular classification benchmark suite, PMLBmini, includes more than 6 times more small-sized datasets than AMLB, OpenML-CC18, and TabZilla combined (Sect.~\ref{sec:relatedwork}). To demonstrate that our dataset selection is not just large in quantity, but also represents a diverse set of data science problems like PMLB \citep{olson2017pmlb,romano2022pmlb}, we summarize key dataset characteristics in Table~\ref{table:characteristics} and plot the feature set size against the sample size for each included dataset in Fig.~\ref{fig:scatter} in Appendix~\ref{sec:appendix_dataset}, showing that our benchmark suite indeed covers a wide range of problem instances with a focus on smaller feature set and sample sizes.

\subsection{Available Methods} \label{sec:methods}

Next to our preselected collection of small-sized tabular datasets, our suite also provides researchers and practitioners with a number of baseline methods. The baselines were chosen to represent a range of different machine learning approaches, but our suite can also be easily extended to include additional pipelines (Sect.~\ref{sec:interface}). We integrated a simple L2-regularized logistic regression classifier \citep{knauer2023cost}, state-of-the-art automated machine learning (AutoML) frameworks \citep{alaa2018autoprognosis,imrie2023autoprognosis,erickson2020autogluon,salinas2023tabrepo}, and recent off-the-shelf deep neural networks \citep{hollmann2023tabpfn,bonet2023hyperfast}. Some of these methods have already been shown to perform well when data is scarce \citep{christodoulou2019systematic,knauer2023cost,mcelfresh2023neural}. An extensive comparison for sample sizes $\leq$ 500 is currently missing, though \citep{bonet2023hyperfast,hollmann2023tabpfn,muller2023mothernet,puri2023semi}. We offer the first comprehensive evaluation for these methods in the low-data regime in Sect.~\ref{sec:results} and assess when state-of-the-art AutoML and deep learning approaches are better suited than a simple logistic regression baseline in this setting in Sect.~\ref{sec:metafeature}.

\myparagraph{Logistic regression} We integrated an L2-regularized logistic regression classifier as a simple, transparent, intrinsically interpretable baseline. We use a continuous conic formulation that is designed to be run-to-run deterministic \citep{aps2023mosek} and can be easily extended to include cardinality or budget constraints for best subset selection \citep{deza2022safe,knauer2023cost}. We re-encode all labels to \{-1, 1\} and ”min-max” scale all features for logistic regression. The L2-regularization hyperparameter $\lambda$ is tuned via a (nested) stratified, 3-fold cross-validation using [0.5, 0.1, 0.02, 0.004] as the hyperparameter grid and the deviance as the validation score.

\myparagraph{AutoML} We focused on 2 recently updated AutoML frameworks for implementation into our suite, AutoPrognosis \citep{alaa2018autoprognosis,imrie2023autoprognosis} and AutoGluon \citep{erickson2020autogluon,salinas2023tabrepo}. AutoPrognosis considers logistic regression and decision tree ensembles to automatically build end-to-end machine learning pipelines, including
preprocessing, model, and hyperparameter selection. Meta-learned hyperparameters from external datasets are used to initialize the default pipeline optimization procedure, and a pipeline ensemble is constructed following the search process. AutoGluon, on the other hand, considers k-nearest neighbors, decision tree ensembles, and neural networks for its algorithm search. It meta-learns a model hyperparameter portfolio from external datasets and different random seeds (zero-shot hyperparameter optimization); the training time budget is then spent on ensembling rather than further hyperparameter optimization. Unfortunately, we had to exclude another state-of-the-art framework, Auto-sklearn \citep{feurer2015efficient,feurer2022auto}, in its current version because the runtime budget was not respected with our setup (Sect.~\ref{sec:setup}) \footnote{\url{https://github.com/automl/auto-sklearn/issues/1683}} \footnote{\url{https://github.com/automl/auto-sklearn/issues/1695}}, but we hope to extend our results in the future.

\myparagraph{Deep learning} We also integrated 2 recent pretrained deep learning models, sometimes referred to as tabular foundation models \citep{muller2023mothernet}, TabPFN \citep{hollmann2023tabpfn} and HyperFast \citep{bonet2023hyperfast}. TabPFN is a meta-trained transformer
ensemble that performs in-context learning on tabular datasets with up to 100 features, without needing any hyperparameter tuning. On our only benchmark dataset with > 100 features, \textit{clean1}, we use subsampling to select 100 features at random for TabPFN \citep{feuer2023scaling}. In contrast to TabPFN, HyperFast uses external datasets to meta-train a hypernetwork, generates smaller, task-specific main networks with the pretrained hypernetwork and the actual training dataset, and optionally fine-tunes and ensembles the main networks.

\vspace{2mm}\noindent With the selected datasets and integrated baselines, we can perform thorough benchmark tests in the low-data regime and analyze when certain approaches succeed or fail, as described in the next section.

\subsection{Python Interface} \label{sec:interface}

PMLBmini is hosted as a Python package on GitHub \footnote{\url{https://github.com/RicardoKnauer/TabMini}}. It can either be imported into an existing project and run in an existing environment, or used standalone in a Docker container. We provide additional information, including how to evaluate a custom tabular classifier on our dataset collection, both on GitHub and in Appendix~\ref{sec:appendix_example}.

\myparagraph{Extensibility} Each baseline classifier (Sect.~\ref{sec:methods}) has been re-implemented as a scikit-learn \texttt{BaseEstimator} with a \texttt{ClassifierMixin}. Therefore, any class that adheres to scikit-learn's standardized duck-typing approach for creating estimators can be used with our benchmarking tool, allowing researchers and practitioners to add new classifiers with minimal overhead. To provide users with a template on how to extend our suite, we implemented logistic regression with a \texttt{Pipeline} and \texttt{GridSearchCV} from scikit-learn.

\myparagraph{The \texttt{tabmini} module}
The benchmarking interface is exposed through the \texttt{tabmini} module, which provides a set of convenience functions for automating the benchmarking process:
\begin{itemize}
    \item \textbf{Benchmark dataset loading} to load our dataset collection (Sect.~\ref{sec:datasets}), with the option to set \texttt{reduced=True} for only loading datasets that have not been used to develop TabPFN's prior (Sect.~\ref{sec:results}).
    The data is returned through a generator. Although this process could have been integrated within the \texttt{compare} function, the latter is able to accept any sort of iterable that maps a string (dataset name) to a tuple of \texttt{pandas dataframes} - design matrix X and labels y. Therefore, the functions have been decoupled.
    \item \textbf{Dummy dataset loading} to validate estimator functionality.
    \item \textbf{Empirical comparison} of a given estimator with the baseline classifiers (Sect.~\ref{sec:results}). By default, we evaluate estimators with the area under the receiver operating characteristic curve (AUC), using a stratified, 3-fold cross validation procedure to obtain mean test performances (Sect.~\ref{sec:setup}). The results of the \texttt{compare} function are returned as a tuple of \texttt{pandas dataframes} - the first dataframe holds the training scores per dataset and estimator, while the second holds test scores.
    \item \textbf{Meta-feature analysis} to extract meta-features from our benchmark datasets and compute associations of these meta-features with performance differences between a given estimator and a simple logistic regression baseline \footnote{\url{https://github.com/RicardoKnauer/TabMini/blob/master/tabmini/analysis/meta\_feature.py}}, allowing us to analyze under which conditions the estimator succeeds or fails in the low-data regime (Sect.~\ref{sec:metafeature}). We use PyMFE 0.4.3 to extract meta-features for all available meta-feature groups \citep{alcobacca2020mfe}, including but not limited to simple meta-features (such as the sample size), statistical meta-features (such as the feature means), clustering meta-features (such as the mean silhouette value), information-theoretic meta-features (such as the feature entropies), model-based meta-features (such as the feature importances in a decision tree), and complexity and landmarking meta-features (such as the predictive performance of a k-nearest neighbors classifier). It is possible for some meta-features, such as the feature importances, to contain multiple values, hence we use all available PyMFE summary functions to aggregate them, including but not limited to the mean and frequency in a particular histogram bin. This yields 3932 meta-features in total for in-depth analyses.
\end{itemize}

\section{Experiments} \label{sec:experiments}

In this section, we describe the experimental setup of our tabular classification benchmark  for data-scarce applications (Sect.~\ref{sec:datasets}) on AutoML and deep learning against logistic regression (Sect.~\ref{sec:methods}) using our suite (Sect.~\ref{sec:interface}). We then present the experimental results, the best L2-regularization hyperparameter for each benchmark dataset, and the conditions under which AutoML frameworks and pretrained deep neural networks outperform logistic regression in the low-data regime. Note that we also provide additional experimental results for gradient-boosted decision trees using their package defaults in Table~\ref{table:results_gbdt} in Appendix~\ref{sec:appendix_results}. Overall, we find that logistic regression shows a similar discriminative performance to AutoML and deep learning approaches on 55\% of the datasets. AutoML methods are better suited for more complex datasets, when more complex classifiers are needed; off-the-shelf deep neural networks are better suited for certain feature distributions, potentially those that resemble their meta-training data.

\subsection{Experimental Setup} \label{sec:setup}

\myparagraph{Method details} We used our logistic regression implementation with the default optimality gaps via MOSEK 10 \citep{aps2023mosek} and JuMP 1.4.0 \citep{dunning2017jump} from Julia 1.8.3. AutoPrognosis 0.1.21 was run from Python 3.10.13 with the default settings, AutoGluon 1.0.0 with the ”best quality” preset, and TabPFN 0.1.9 with 32 ensemble members \citep{hollmann2023tabpfn}. HyperFast 0.1.3 was used without fine-tuning of the main networks (due to excessive runtimes exceeding the time budget on our hardware configuration, see below), but with 32 ensemble members like TabPFN \citep{bonet2023hyperfast,hollmann2023tabpfn}.

\myparagraph{Evaluation metrics} We measured the discriminative performance in terms of the AUC. The training AUC was recorded to assess overfitting and evaluated with a 1h runtime limit for each method per benchmark dataset, the mean test AUC with a 3h runtime limit via a stratified, 3-fold cross-validation procedure (i.e., 1h per fold). The time budget was chosen to reflect current practice and to ensure comparability with prior work \citep{bonet2023hyperfast,erickson2020autogluon,hollmann2023tabpfn,muller2023mothernet,salinas2023tabrepo}. For logistic regression, we also tracked the best L2-regularization hyperparameter value $\lambda^*$ obtained on the whole data, and used a nested instead of a non-nested cross-validation procedure \citep{knauer2023cost}. To detect pairwise mean test AUC differences between our methods, we used a critical difference diagram based on the Holm-adjusted Wilcoxon signed-rank test with a 0.05 significance level \citep{demsar2006statistical,benavoli2016should}. Each experiment was run on 8 vCPU cores with 32GiB memory on an internal cluster.

\subsection{Experimental Results} \label{sec:results}

\begin{figure}[t]
\centering
\begin{subfigure}{0.64\textwidth}
\includegraphics[width=\textwidth]{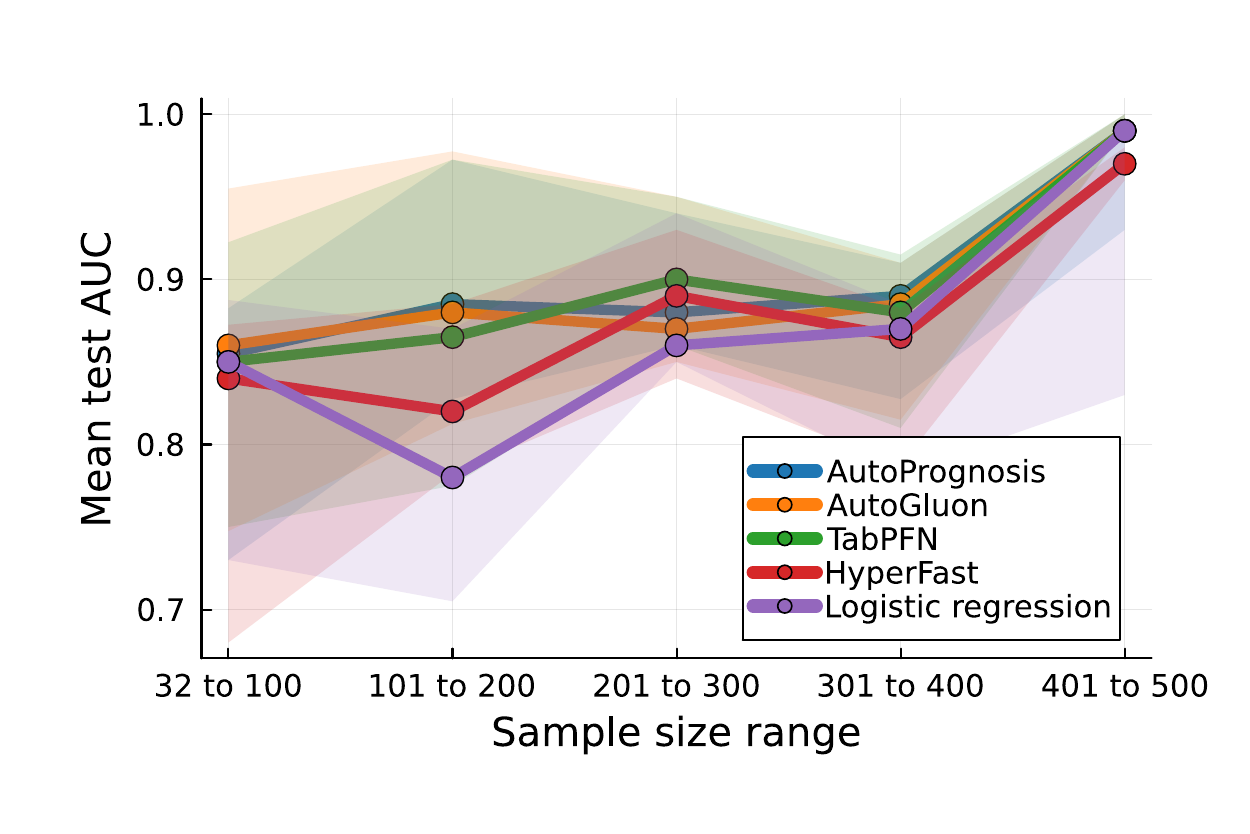}
\caption{Mean test AUC medians and interquartile ranges across sample size ranges in steps of 100.}
\label{fig:results_a}
\end{subfigure}
\hspace{2mm}
\begin{subfigure}{0.33\textwidth}
\includegraphics[width=\textwidth]{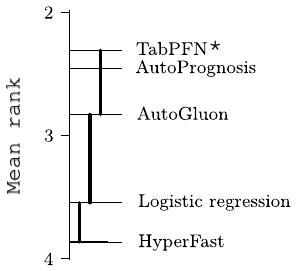}
\caption{Critical difference diagram, with a bold vertical bar connecting methods that are not statistically different. $^\mathbf{\star}$Note that TabPFN results are biased (Sect.~\ref{sec:results}).}
\label{fig:results_b}
\end{subfigure}
\caption{Discriminative performance for AutoML, deep learning, and logistic regression on our benchmark suite PMLBmini.}
\label{fig:results_ab}
\end{figure}

AutoPrognosis, AutoGluon, TabPFN, HyperFast, and logistic regression show a relatively similar discriminative performance at different sample sizes (Fig.~\ref{fig:results_a}). The largest difference occurs at sample sizes from 101 to 200, with HyperFast and logistic regression only reaching a mean test AUC of 0.82 and 0.78, respectively. There is a trend for all methods to perform better when the sample size increases (Fig.~\ref{fig:results_a}). Each approach wins on at least one benchmark dataset and looses on at least one other dataset (Table~\ref{table:results} in Appendix~\ref{sec:appendix_results}), echoing the results of \citet{mcelfresh2023neural}. Logistic regression performs on par with or better than the best AutoML or deep learning approach on 16\% of the datasets, and lies within 1\%, 2\%, and 3\% of the best approach on 34\%, 48\%, and 55\% of the datasets, respectively - possibly because it is less likely to overfit, especially with smaller sample sizes. Interestingly, even when AutoML methods consider logistic regression for algorithm selection (i.e., AutoPrognosis), they may be outperformed by logistic regression alone (e.g., on the backache dataset), again possibly due to overfitting. When statistically comparing pairwise mean test AUC differences (Fig.~\ref{fig:results_b}), we observe that AutoPrognosis and TabPFN achieve a better rank than logistic regression (p $<$ 0.05), whereas AutoGluon and HyperFast are not different from a simple logistic regression baseline (p > 0.05). TabPFN's prior was developed on 45\% of our benchmark datasets, though \citep{hollmann2023tabpfn}, its performance estimates are therefore likely to be overoptimistic. However, manually excluding all overlapping datasets reduced the benchmark dataset collection in our analysis too much to find statistically significant performance differences between any of the evaluated methods (p > 0.05). Finally, we report the best L2-regularization hyperparameter for each dataset that can be used for meta-learning (Table~\ref{table:results} in Appendix~\ref{sec:appendix_results}), for instance by leveraging the most similar PMLBmini dataset(s) for zero- or few-shot hyperparameter optimization \citep{feurer2015initializing,reif2012meta}.

\subsection{Meta-Feature Analysis} \label{sec:metafeature}

\begin{figure}[t]
\centering
\includegraphics[width=0.7\textwidth]{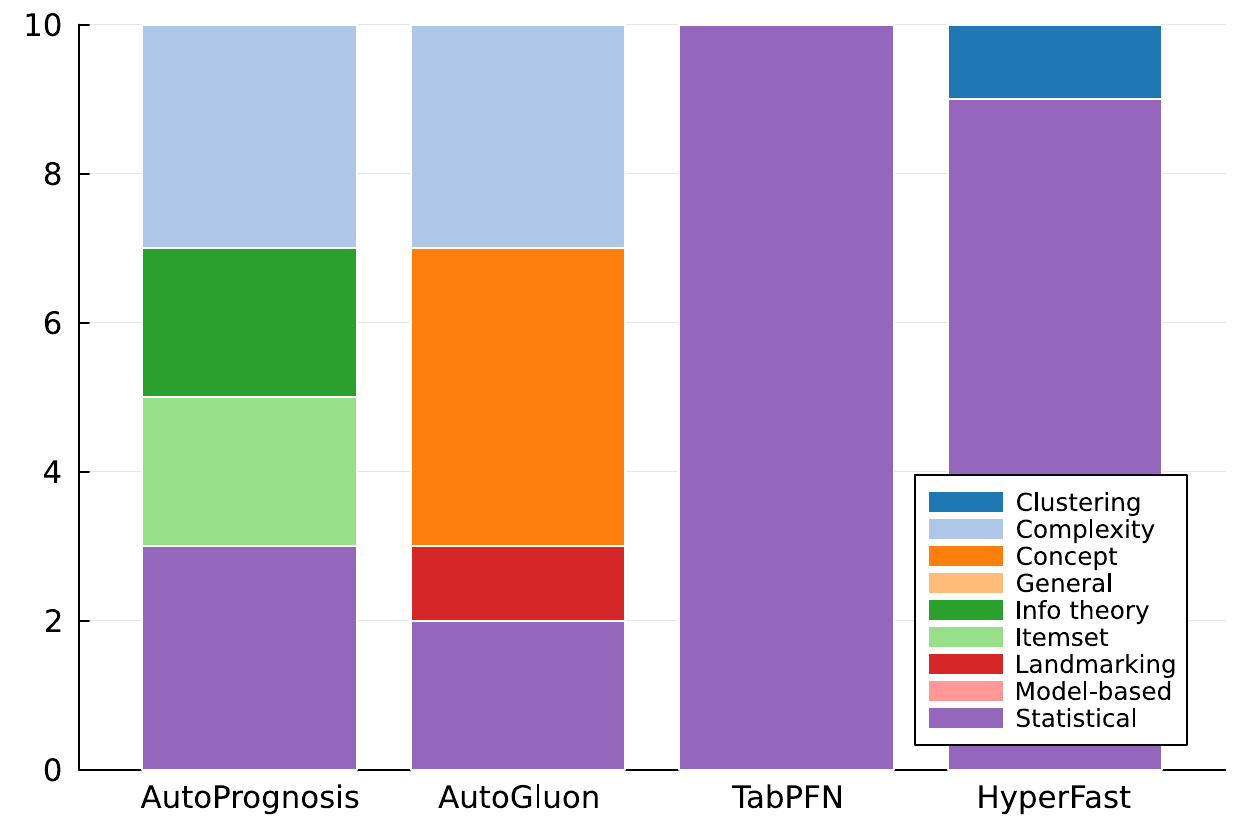}
\caption{What dataset meta-features influence model performance? The plot shows the meta-feature groups \citep{alcobacca2020mfe} that are represented in the top-10 meta-features per approach. To that end, we computed all PyMFE meta-features per dataset, the mean test AUC differences between each AutoML / deep learning method and logistic regression per dataset, the absolute Spearman rank correlation coefficient between each PyMFE meta-feature and the performance difference across datasets (Sect.~\ref{sec:interface}); and finally selected the top-10 meta-features with the largest absolute correlations.}
\label{fig:results_c}
\end{figure}

In the following, we use our meta-feature analysis tool to extract meta-features from our benchmark datasets and compute relationships of these meta-features with performance differences between each AutoML / deep learning method and logistic regression (Sect.~\ref{sec:interface}). This way, we can determine which dataset properties make more complex machine learning methods well- or less well-suited than a simple logistic regression baseline in the low-data regime. Fig.~\ref{fig:results_d} in Appendix~\ref{sec:appendix_results} shows the top-3 correlations for each approach. For AutoPrognosis, the most discriminating meta-feature is the ratio of the intra- and extraclass nearest neighbor distance; for AutoGluon, it is the clustering coeffcient. Both measures capture the dataset complexity, are larger for harder classification problems \citep{lorena2019complex}, and show a positive relationship with the performance differences. For AutoML, most meta-features with the largest absolute correlations in fact measure the dataset complexity (Fig.~\ref{fig:results_c}). For TabPFN and HyperFast, the most discriminating meta-features are the harmonic mean and minimum of each feature, i.e., summary scores from the feature distribution. For deep learning, meta-features with the largest absolute correlations are almost exclusively statistical meta-features that describe the feature distribution (Fig.~\ref{fig:results_c}). Therefore, AutoML methods appear to be better suited for more complex datasets that need more complex classifiers, whereas pretrained deep neural networks may be better suited for datasets with feature distributions that resemble their meta-training data. These insights may prove beneficial when developing new or updating existing machine learning systems, for example by increasing the meta-training set diversity for off-the-shelf deep learning models in the low-data regime.

\section{Broader Impact and Limitations} \label{sec:broader}

Our benchmarking tool, PMLBmini, provides a standardized, diverse collection of 44 small-sized tabular datasets (Sect.~\ref{sec:datasets}) in combination with a range of different machine learning baselines (Sect.~\ref{sec:methods}), accessible via a user-friendly Python interface (Sect.~\ref{sec:interface}). This allows researchers and practitioners working with tabular data to rigorously challenge their own classifier for data-scarce applications without much effort, and the community to more easily track progress in the field. Interestingly, our initial set of machine learning approaches reveals that simple baselines like logistic regression should not be prematurely discarded since they frequently perform similar to state-of-the-art AutoML and deep learning methods when data is limited (Sect.~\ref{sec:results}). In fact, trying a simple logistic regression baseline in the low-data regime first and only switching to more complex approaches when needed could save resources \citep{mcelfresh2023neural} and improve the transparency, trust, and applicability of machine learning systems in practice, for example in the clinical domain \citep{falla2021machine,knauer2023cost,steyerberg2019clinical}. We also provide users with a practical meta-feature analysis tool to investigate under which conditions, i.e., dataset properties, certain methods are better suited than others. AutoML frameworks appear to work better when more complex classifiers are needed, pretrained deep learning models when their meta-training data are more similar to the test data (Sect.~\ref{sec:metafeature}). We hope that this will support researchers and practitioners to develop new or improve upon existing machine learning systems, for instance by meta-training deep neural networks on a wider range of small-sized tabular data.

Nevertheless, the development of a benchmark suite also carries risks. As our selected datasets are publicly available, we cannot guarantee that they have not already been used for training or tuning the system that is intended to be benchmarked. As many state-of-the-art AutoML and deep learning methods leverage meta-learning (Sect.~\ref{sec:methods}), it is quite possible that meta-training and benchmark datasets overlap to a large extent for some approaches - in this case, users could either accept the inherent bias in the benchmark results or have the option to manually exclude the affected datasets (Sect.~\ref{sec:interface}). Overoptimism and "arbitrary" dataset exclusions can potentially undermine the suite's promise to increase the comparability between methods and studies, though. Moreover, we also want to emphasize that our benchmark suite only contains binary classification problems without missing values and with categorical features being numerically encoded (Sect.~\ref{sec:datasets}), i.e., categorical features contain no more textual information that could be used by AutoML methods \citep{erickson2020autogluon} or multimodal foundation models \citep{achiam2023gpt}. For now, PMLBmini therefore only represents a small slice of the data-scarce problems commonly encountered in practice and disadvantages approaches that naturally handle missing values and categorical features such as gradient-boosted decision trees. Additionally, our meta-feature analysis tool currently only relies on a bivariate association measure that does not take confounding factors into account and is not designed to detect meta-feature interactions or non-monotonic relationships \citep{gijsbers2022amlb,mcelfresh2023neural}. Further limitations could be addressed by integrating assessments for training or inference speed and encountered errors \citep{gijsbers2022amlb,mcelfresh2023neural}, scaling to slightly larger sample sizes for finding out at which point more complex methods start to consistently outperform simple baselines, and subsampling larger tabular datasets to increase the number of benchmark datasets in our suite. We strongly encourage future research and contributions in these directions.

\section{Conclusion}

Although researchers and practitioners are frequently confronted with data-scarce applications, (very-)small sized tabular data are generally underrepresented in current machine learning benchmarks. In this work, we introduced PMLBmini, a tabular benchmark suite of 44 binary classification datasets with sample sizes $\leq$ 500. We showed how our benchmarking tool can be used to evaluate current AutoML and deep learning methods, and to analyze if and when they are better suited than a simple logistic regression baseline. In summary, we found that state-of-the-art machine learning approaches fail to appreciably outperform logistic regression on 55\% of our benchmark datasets. AutoML frameworks appear to work better for more complex datasets, pretrained deep neural networks for datasets with feature distributions that are more similar to their meta-training data. We therefore recommend to increase the dataset diversity when meta-training off-the-shelf deep neural networks. Since hyperparameter optimization in the low-data regime is inherently difficult, we also provide the community with L2-regularization hyperparamters that can be directly used for meta-learning when data is limited. Finally, we encourage researchers and practitioners to challenge the data efficiency of their own tabular classifier using our suite, and to assess under which conditions it succeeds or fails.

\begin{acknowledgements}
This research was funded by the Bundesministerium für Bildung und Forschung (16DHBKI071, 01IS23041C).
\end{acknowledgements}

\bibliography{automl.bib}

\begin{thebibliography}{}

\bibitem[Achiam et~al., 2023]{achiam2023gpt}
Achiam, J., Adler, S., Agarwal, S., Ahmad, L., Akkaya, I., Aleman, F.~L., Almeida, D., Altenschmidt, J., Altman, S., Anadkat, S., et~al. (2023).
\newblock {GPT}-4 technical report.
\newblock {\em arXiv preprint arXiv:2303.08774}.

\bibitem[Alaa and van~der Schaar, 2018]{alaa2018autoprognosis}
Alaa, A. and van~der Schaar, M. (2018).
\newblock Autoprognosis: Automated clinical prognostic modeling via bayesian optimization with structured kernel learning.
\newblock In {\em International Conference on Machine Learning (ICML)}, pages 139--148. PMLR.

\bibitem[Alcoba{\c{c}}a et~al., 2020]{alcobacca2020mfe}
Alcoba{\c{c}}a, E., Siqueira, F., Rivolli, A., Garcia, L.~P., Oliva, J.~T., and De~Carvalho, A.~C. (2020).
\newblock Mfe: Towards reproducible meta-feature extraction.
\newblock {\em The Journal of Machine Learning Research}, 21(1):4503--4507.

\bibitem[Benavoli et~al., 2016]{benavoli2016should}
Benavoli, A., Corani, G., and Mangili, F. (2016).
\newblock Should we really use post-hoc tests based on mean-ranks?
\newblock {\em The Journal of Machine Learning Research}, 17(1):152--161.

\bibitem[Betker et~al., 2023]{betker2023improving}
Betker, J., Goh, G., Jing, L., Brooks, T., Wang, J., Li, L., Ouyang, L., Zhuang, J., Lee, J., Guo, Y., et~al. (2023).
\newblock Improving image generation with better captions.
\newblock {\em Computer Science. https://cdn. openai. com/papers/dall-e-3. pdf}, 2(3):8.

\bibitem[Bischl et~al., 2021]{bischl2021openml}
Bischl, B., Casalicchio, G., Feurer, M., Gijsbers, P., Hutter, F., Lang, M., Mantovani, R.~G., van Rijn, J.~N., and Vanschoren, J. (2021).
\newblock {OpenML} benchmarking suites.
\newblock In {\em Proceedings of the NeurIPS 2021 Datasets and Benchmarks Track}.

\bibitem[Bonet et~al., 2024]{bonet2023hyperfast}
Bonet, D., Montserrat, D.~M., Gir{\'o}-i Nieto, X., and Ioannidis, A.~G. (2024).
\newblock Hyperfast: Instant classification for tabular data.
\newblock {\em arXiv preprint arXiv:2402.14335}.

\bibitem[Borisov et~al., 2022]{borisov2022deep}
Borisov, V., Leemann, T., Se{\ss}ler, K., Haug, J., Pawelczyk, M., and Kasneci, G. (2022).
\newblock Deep neural networks and tabular data: A survey.
\newblock {\em IEEE Transactions on Neural Networks and Learning Systems}.

\bibitem[Brooks et~al., 2024]{videoworldsimulators2024}
Brooks, T., Peebles, B., Homes, C., DePue, W., Guo, Y., Jing, L., Schnurr, D., Taylor, J., Luhman, T., Luhman, E., Ng, C. W.~Y., Wang, R., and Ramesh, A. (2024).
\newblock Video generation models as world simulators.

\bibitem[Brown et~al., 2020]{brown2020language}
Brown, T., Mann, B., Ryder, N., Subbiah, M., Kaplan, J.~D., Dhariwal, P., Neelakantan, A., Shyam, P., Sastry, G., Askell, A., et~al. (2020).
\newblock Language models are few-shot learners.
\newblock {\em Advances in neural information processing systems}, 33:1877--1901.

\bibitem[Christodoulou et~al., 2019]{christodoulou2019systematic}
Christodoulou, E., Ma, J., Collins, G.~S., Steyerberg, E.~W., Verbakel, J.~Y., and Van~Calster, B. (2019).
\newblock A systematic review shows no performance benefit of machine learning over logistic regression for clinical prediction models.
\newblock {\em Journal of clinical epidemiology}, 110:12--22.

\bibitem[Dem{\v{s}}ar, 2006]{demsar2006statistical}
Dem{\v{s}}ar, J. (2006).
\newblock Statistical comparisons of classifiers over multiple data sets.
\newblock {\em The Journal of Machine learning research}, 7(1):1--30.

\bibitem[Deza and Atamt{\"u}rk, 2022]{deza2022safe}
Deza, A. and Atamt{\"u}rk, A. (2022).
\newblock Safe screening for logistic regression with l0-l2 regularization.
\newblock {\em arXiv}, 2202.

\bibitem[Dunning et~al., 2017]{dunning2017jump}
Dunning, I., Huchette, J., and Lubin, M. (2017).
\newblock Jump: A modeling language for mathematical optimization.
\newblock {\em SIAM review}, 59(2):295--320.

\bibitem[Erickson et~al., 2020]{erickson2020autogluon}
Erickson, N., Mueller, J., Shirkov, A., Zhang, H., Larroy, P., Li, M., and Smola, A. (2020).
\newblock Autogluon-tabular: Robust and accurate automl for structured data.
\newblock {\em arXiv preprint arXiv:2003.06505}.

\bibitem[Falla et~al., 2021]{falla2021machine}
Falla, D., Devecchi, V., Jim{\'e}nez-Grande, D., R{\"u}gamer, D., and Liew, B.~X. (2021).
\newblock Machine learning approaches applied in spinal pain research.
\newblock {\em Journal of Electromyography and Kinesiology}, 61:102599.

\bibitem[Feuer et~al., 2023]{feuer2023scaling}
Feuer, B., Hegde, C., and Cohen, N. (2023).
\newblock Scaling tabpfn: Sketching and feature selection for tabular prior-data fitted networks.
\newblock In {\em NeurIPS 2023 Second Table Representation Learning Workshop}.

\bibitem[Feurer et~al., 2022]{feurer2022auto}
Feurer, M., Eggensperger, K., Falkner, S., Lindauer, M., and Hutter, F. (2022).
\newblock Auto-sklearn 2.0: Hands-free automl via meta-learning.
\newblock {\em Journal of Machine Learning Research}, 23(261):1--61.

\bibitem[Feurer et~al., 2015a]{feurer2015efficient}
Feurer, M., Klein, A., Eggensperger, K., Springenberg, J., Blum, M., and Hutter, F. (2015a).
\newblock Efficient and robust automated machine learning.
\newblock {\em Advances in neural information processing systems}, 28.

\bibitem[Feurer et~al., 2015b]{feurer2015initializing}
Feurer, M., Springenberg, J., and Hutter, F. (2015b).
\newblock Initializing bayesian hyperparameter optimization via meta-learning.
\newblock In {\em Proceedings of the AAAI Conference on Artificial Intelligence}, volume~29.

\bibitem[Fischer et~al., 2023]{fischer2023openml}
Fischer, S.~F., Feurer, M., and Bischl, B. (2023).
\newblock {OpenML-CTR23}--a curated tabular regression benchmarking suite.
\newblock In {\em AutoML Conference 2023 (Workshop)}.

\bibitem[Gijsbers et~al., 2022]{gijsbers2022amlb}
Gijsbers, P., Bueno, M.~L., Coors, S., LeDell, E., Poirier, S., Thomas, J., Bischl, B., and Vanschoren, J. (2022).
\newblock {AMLB}: an {AutoML} benchmark.
\newblock {\em arXiv preprint arXiv:2207.12560}.

\bibitem[Gijsbers et~al., 2019]{gijsbers2019open}
Gijsbers, P., LeDell, E., Thomas, J., Poirier, S., Bischl, B., and Vanschoren, J. (2019).
\newblock An open source {AutoML} benchmark.
\newblock {\em arXiv preprint arXiv:1907.00909}.

\bibitem[Hollmann et~al., 2023]{hollmann2023tabpfn}
Hollmann, N., M{\"u}ller, S., Eggensperger, K., and Hutter, F. (2023).
\newblock Tabpfn: A transformer that solves small tabular classification problems in a second.
\newblock In {\em Proceedings of the International Conference on Learning Representations (ICLR)}.

\bibitem[Hulsebos et~al., 2023]{hulsebos2023gittables}
Hulsebos, M., Demiralp, {\c{C}}., and Groth, P. (2023).
\newblock Gittables: A large-scale corpus of relational tables.
\newblock {\em Proceedings of the ACM on Management of Data}, 1(1):1--17.

\bibitem[Imrie et~al., 2023]{imrie2023autoprognosis}
Imrie, F., Cebere, B., McKinney, E.~F., and van~der Schaar, M. (2023).
\newblock Autoprognosis 2.0: Democratizing diagnostic and prognostic modeling in healthcare with automated machine learning.
\newblock {\em PLOS Digital Health}, 2(6):e0000276.

\bibitem[Knauer and Rodner, 2023]{knauer2023cost}
Knauer, R. and Rodner, E. (2023).
\newblock Cost-sensitive best subset selection for logistic regression: A mixed-integer conic optimization perspective.
\newblock In {\em German Conference on Artificial Intelligence (K{\"u}nstliche Intelligenz)}, pages 114--129. Springer.

\bibitem[Lorena et~al., 2019]{lorena2019complex}
Lorena, A.~C., Garcia, L.~P., Lehmann, J., Souto, M.~C., and Ho, T.~K. (2019).
\newblock How complex is your classification problem? a survey on measuring classification complexity.
\newblock {\em ACM Computing Surveys (CSUR)}, 52(5):1--34.

\bibitem[McElfresh et~al., 2024]{mcelfresh2023neural}
McElfresh, D., Khandagale, S., Valverde, J., Prasad~C, V., Ramakrishnan, G., Goldblum, M., and White, C. (2024).
\newblock When do neural nets outperform boosted trees on tabular data?
\newblock {\em Advances in Neural Information Processing Systems}, 36.

\bibitem[Moons et~al., 2015]{moons2015transparent}
Moons, K.~G., Altman, D.~G., Reitsma, J.~B., Ioannidis, J.~P., Macaskill, P., Steyerberg, E.~W., Vickers, A.~J., Ransohoff, D.~F., and Collins, G.~S. (2015).
\newblock Transparent reporting of a multivariable prediction model for individual prognosis or diagnosis ({TRIPOD}): explanation and elaboration.
\newblock {\em Annals of internal medicine}, 162(1):W1--W73.

\bibitem[Moons et~al., 2019]{moons2019probast}
Moons, K.~G., Wolff, R.~F., Riley, R.~D., Whiting, P.~F., Westwood, M., Collins, G.~S., Reitsma, J.~B., Kleijnen, J., and Mallett, S. (2019).
\newblock {PROBAST}: a tool to assess risk of bias and applicability of prediction model studies: explanation and elaboration.
\newblock {\em Annals of internal medicine}, 170(1):W1--W33.

\bibitem[{MOSEK ApS}, 2023]{aps2023mosek}
{MOSEK ApS} (2023).
\newblock {MOSEK} optimizer {API} for {Python}.
\newblock Manual.

\bibitem[M{\"u}ller et~al., 2023]{muller2023mothernet}
M{\"u}ller, A., Curino, C., and Ramakrishnan, R. (2023).
\newblock Mothernet: A foundational hypernetwork for tabular classification.
\newblock {\em arXiv preprint arXiv:2312.08598}.

\bibitem[Olson et~al., 2017]{olson2017pmlb}
Olson, R.~S., La~Cava, W., Orzechowski, P., Urbanowicz, R.~J., and Moore, J.~H. (2017).
\newblock {PMLB}: a large benchmark suite for machine learning evaluation and comparison.
\newblock {\em BioData mining}, 10:1--13.

\bibitem[Puri et~al., 2023]{puri2023semi}
Puri, V., Kataria, A., and Puri, B. (2023).
\newblock Semi-automated diabetes prediction using autogluon and tabpfn models.
\newblock In {\em International Conference on Artificial Intelligence of Things}, pages 289--295. Springer.

\bibitem[Reif et~al., 2012]{reif2012meta}
Reif, M., Shafait, F., and Dengel, A. (2012).
\newblock Meta-learning for evolutionary parameter optimization of classifiers.
\newblock {\em Machine learning}, 87:357--380.

\bibitem[Riley et~al., 2021]{riley2021penalization}
Riley, R.~D., Snell, K.~I., Martin, G.~P., Whittle, R., Archer, L., Sperrin, M., and Collins, G.~S. (2021).
\newblock Penalization and shrinkage methods produced unreliable clinical prediction models especially when sample size was small.
\newblock {\em Journal of Clinical Epidemiology}, 132:88--96.

\bibitem[Romano et~al., 2022]{romano2022pmlb}
Romano, J.~D., Le, T.~T., La~Cava, W., Gregg, J.~T., Goldberg, D.~J., Chakraborty, P., Ray, N.~L., Himmelstein, D., Fu, W., and Moore, J.~H. (2022).
\newblock {PMLB} v1. 0: an open-source dataset collection for benchmarking machine learning methods.
\newblock {\em Bioinformatics}, 38(3):878--880.

\bibitem[Salinas and Erickson, 2023]{salinas2023tabrepo}
Salinas, D. and Erickson, N. (2023).
\newblock Tabrepo: A large scale repository of tabular model evaluations and its automl applications.
\newblock {\em arXiv preprint arXiv:2311.02971}.

\bibitem[Shwartz-Ziv and Armon, 2022]{shwartz2022tabular}
Shwartz-Ziv, R. and Armon, A. (2022).
\newblock Tabular data: Deep learning is not all you need.
\newblock {\em Information Fusion}, 81:84--90.

\bibitem[{\v{S}}inkovec et~al., 2021]{vsinkovec2021tune}
{\v{S}}inkovec, H., Heinze, G., Blagus, R., and Geroldinger, A. (2021).
\newblock To tune or not to tune, a case study of ridge logistic regression in small or sparse datasets.
\newblock {\em BMC Medical Research Methodology}, 21:1--15.

\bibitem[Steyerberg, 2019]{steyerberg2019clinical}
Steyerberg, E.~W. (2019).
\newblock {\em Clinical prediction models: a practical approach to development, validation, and updating}.
\newblock Springer.

\bibitem[Van~Calster et~al., 2020]{van2020regression}
Van~Calster, B., van Smeden, M., De~Cock, B., and Steyerberg, E.~W. (2020).
\newblock Regression shrinkage methods for clinical prediction models do not guarantee improved performance: simulation study.
\newblock {\em Statistical methods in medical research}, 29(11):3166--3178.

\bibitem[Vanschoren et~al., 2014]{vanschoren2014openml}
Vanschoren, J., Van~Rijn, J.~N., Bischl, B., and Torgo, L. (2014).
\newblock {OpenML}: networked science in machine learning.
\newblock {\em ACM SIGKDD Explorations Newsletter}, 15(2):49--60.

\end{thebibliography}

\newpage 
\section*{Submission Checklist}

\begin{enumerate}
\item For all authors\dots
  \begin{enumerate}
  \item Do the main claims made in the abstract and introduction accurately
    reflect the paper's contributions and scope?
    \answerYes{We made sure that the claims in the abstract and introduction are coherent with the contributions.}
  \item Did you describe the limitations of your work?
    \answerYes{See Sect.~\ref{sec:broader}.}
  \item Did you discuss any potential negative societal impacts of your work?
    \answerYes{See Sect.~\ref{sec:broader}.}
  \item Did you read the ethics review guidelines and ensure that your paper
    conforms to them? \url{https://2022.automl.cc/ethics-accessibility/}
    \answerYes{See Sect.~\ref{sec:broader}.}
  \end{enumerate}
\item If you ran experiments\dots
  \begin{enumerate}
  \item Did you use the same evaluation protocol for all methods being compared (e.g., 
    same benchmarks, data (sub)sets, available resources)? 
   \answerYes{See Sect.~\ref{sec:experiments}.}
  \item Did you specify all the necessary details of your evaluation (e.g., data splits,
    pre-processing, search spaces, hyperparameter tuning)?
    \answerYes{See Sect.~\ref{sec:methods} and \ref{sec:experiments}.}
  \item Did you repeat your experiments (e.g., across multiple random seeds or splits) to account for the impact of randomness in your methods or data?
    \answerYes{We accounted for randomness by calculating the mean test AUC from 3 folds for each method per benchmark dataset.}
  \item Did you report the uncertainty of your results (e.g., the variance across random seeds or splits)?
    \answerYes{We considered the uncertainty across benchmark datasets, both in terms of interquartile ranges across different sample size ranges and in terms of Holm-adjusted Wilcoxon signed-rank tests to detect pairwise mean test AUC differences between methods.}
  \item Did you report the statistical significance of your results?
    \answerYes{See Sect.~\ref{sec:experiments}.}
  \item Did you use tabular or surrogate benchmarks for in-depth evaluations?
    \answerNo{We propose our own benchmark, and conduct in-depth evaluations on it.}
  \item Did you compare performance over time and describe how you selected the maximum duration?
    \answerNo{We did not compare performance over time, but chose a fixed runtime limit of 4h per benchmark dataset for each method. The details and reasoning are described in Sect.~\ref{sec:setup}. We concede that recording anytime performance can be very insightful for AutoML frameworks, though, to assess how quickly they converge.}
  \item Did you include the total amount of compute and the type of resources
    used (e.g., type of \textsc{gpu}s, internal cluster, or cloud provider)?
    \answerYes{See Sect.~\ref{sec:setup}.}
  \item Did you run ablation studies to assess the impact of different
    components of your approach?
    \answerNo{In our experiments, we were primarily interested in comparing a range of existing machine learning approaches on our benchmark datasets in a plug-and-play fashion, mostly using the default settings. However, PMLBmini allows users to easily implement custom machine learning pipelines (Sect.~\ref{sec:interface}), and we encourage further research into which specific pipeline components yield appreciable performance benefits in the low-data regime.}
  \end{enumerate}
\item With respect to the code used to obtain your results\dots
  \begin{enumerate}
\item Did you include the code, data, and instructions needed to reproduce the
    main experimental results, including all requirements (e.g.,
    \texttt{requirements.txt} with explicit versions), random seeds, an instructive
    \texttt{README} with installation, and execution commands (either in the
    supplemental material or as a \textsc{url})?
    \answerYes{See \url{https://github.com/RicardoKnauer/TabMini}.}
  \item Did you include a minimal example to replicate results on a small subset
    of the experiments or on toy data?
    \answerYes{See \url{https://github.com/RicardoKnauer/TabMini}.}
  \item Did you ensure sufficient code quality and documentation so that someone else 
    can execute and understand your code?
    \answerYes{See \url{https://github.com/RicardoKnauer/TabMini}.}
  \item Did you include the raw results of running your experiments with the given
    code, data, and instructions?
    \answerYes{See \url{https://github.com/RicardoKnauer/TabMini}.}
  \item Did you include the code, additional data, and instructions needed to generate
    the figures and tables in your paper based on the raw results?
    \answerYes{See \url{https://github.com/RicardoKnauer/TabMini}.}
  \end{enumerate}
\item If you used existing assets (e.g., code, data, models)\dots
  \begin{enumerate}
  \item Did you cite the creators of used assets?
    \answerYes{See Sect.~\ref{sec:datasets} and \ref{sec:methods}.}
  \item Did you discuss whether and how consent was obtained from people whose
    data you're using/curating if the license requires it?
    \answerYes{The PMLB license does not require it.}
  \item Did you discuss whether the data you are using/curating contains
    personally identifiable information or offensive content?
    \answerYes{See Sect.~\ref{sec:datasets}.}
  \end{enumerate}
\item If you created/released new assets (e.g., code, data, models)\dots
  \begin{enumerate}
    \item Did you mention the license of the new assets (e.g., as part of your code submission)?
    \answerYes{See \url{https://github.com/RicardoKnauer/TabMini}.}
    \item Did you include the new assets either in the supplemental material or as
    a \textsc{url} (to, e.g., GitHub or Hugging Face)?
    \answerYes{See \url{https://github.com/RicardoKnauer/TabMini}.}
  \end{enumerate}
\item If you used crowdsourcing or conducted research with human subjects\dots
  \begin{enumerate}
  \item Did you include the full text of instructions given to participants and
    screenshots, if applicable?
    \answerNA{We did not use crowdsourcing or conducted research with human subjects.}
  \item Did you describe any potential participant risks, with links to
    Institutional Review Board (\textsc{irb}) approvals, if applicable?
    \answerNA{We did not use crowdsourcing or conducted research with human subjects.}
  \item Did you include the estimated hourly wage paid to participants and the
    total amount spent on participant compensation?
    \answerNA{We did not use crowdsourcing or conducted research with human subjects.}
  \end{enumerate}
\item If you included theoretical results\dots
  \begin{enumerate}
  \item Did you state the full set of assumptions of all theoretical results?
    \answerNA{We did not include theoretical results.}
  \item Did you include complete proofs of all theoretical results?
    \answerNA{We did not include theoretical results.}
  \end{enumerate}
\end{enumerate}

\newpage

\appendix
\section{Additional Dataset Details} \label{sec:appendix_dataset}

\begin{figure}[h]
\centering
\begin{subfigure}{0.53\textwidth}
\includegraphics[width=\textwidth]{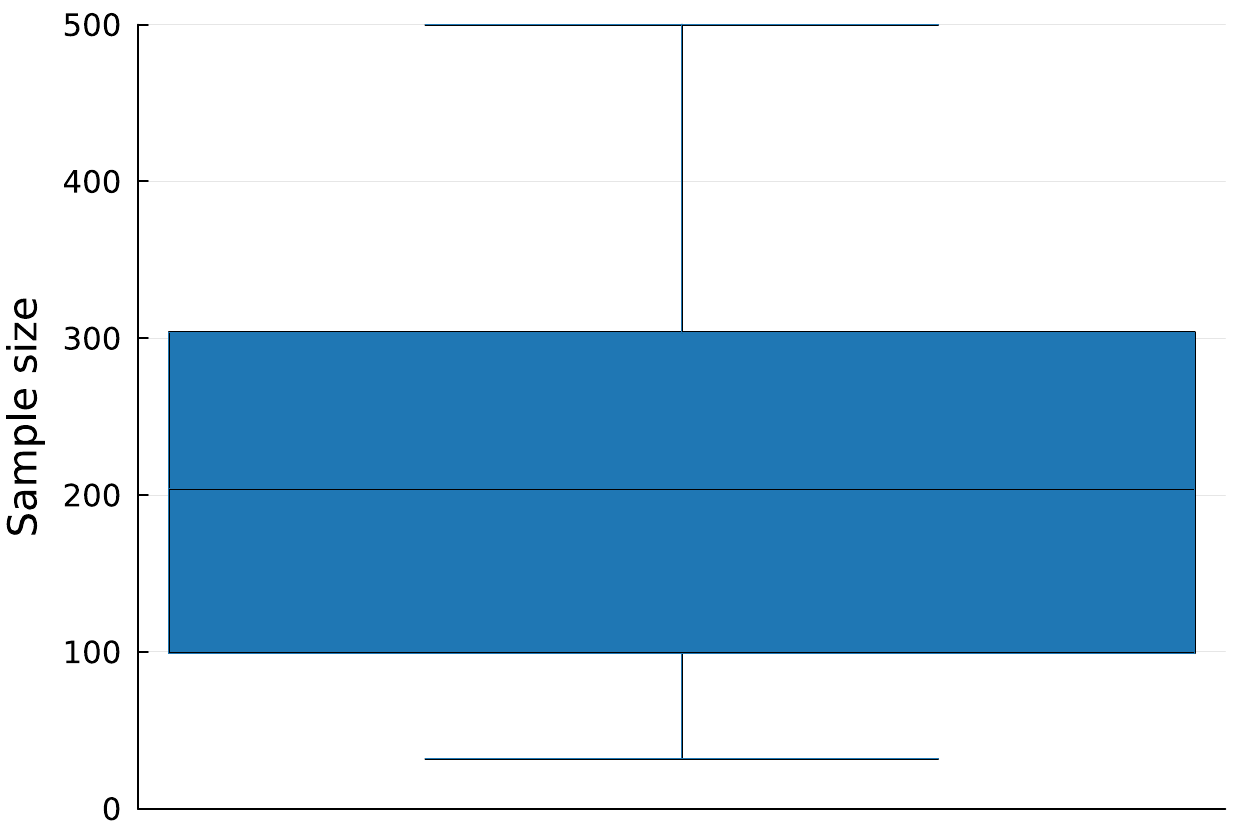}
\caption{Dataset dimensionality in terms of the number of instances across all datasets.}
\label{fig:boxplot}
\end{subfigure}
\hspace{2mm}
\begin{subfigure}{0.44\textwidth}
\includegraphics[width=\textwidth]{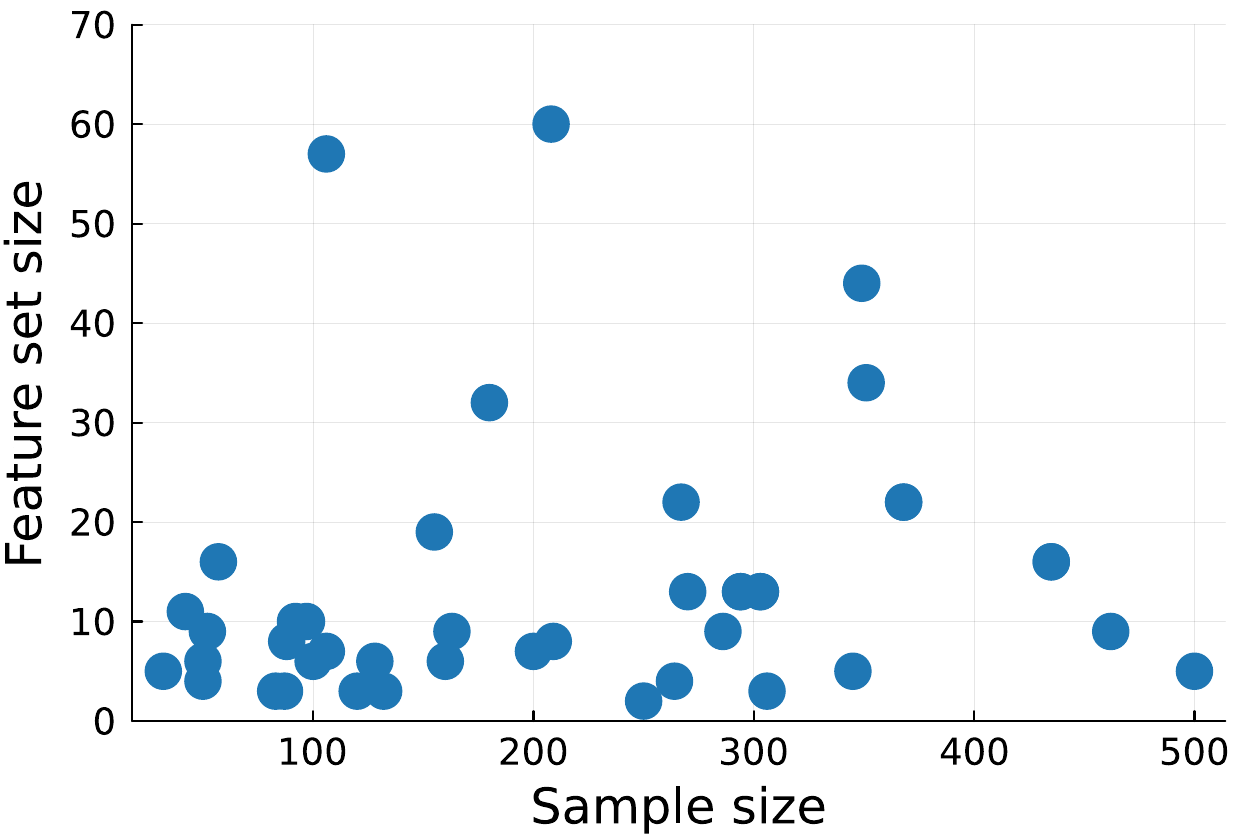}
\caption{Number of features and instances for each dataset. The \textit{clean1} dataset with a feature set size of 168 is not shown to make the plot more readable.}
\label{fig:scatter}
\end{subfigure}
\caption{Dataset dimensionality in our benchmark suite.}
\label{fig:dimensionality}
\end{figure}

In the following, we provide additional details on our dataset collection in terms of key dataset characteristics in Table~\ref{table:characteristics} and in terms of dataset dimensionality in Fig.~\ref{fig:dimensionality}.

\tabcolsep=0.45cm
\begin{table}[h]
  \centering
  \caption{Key summary statistics for our benchmark suite in terms of the sample size, feature set size, relative minority class frequency, events per variable (number of instances in the minority class per feature), and number of binary features.}
  \begin{tabular}{p{1.5cm} p{1.5cm} p{1.5cm} p{1.5cm} p{1.7cm} p{1.5cm}}
  \\ \hline \\
    \textbf{Summary statistic} & \textbf{Sample size} & \textbf{Feature set size} & \textbf{\% Minority class} & \textbf{Events per variable} & \textbf{\# Binary features}
  \\ \hline \\
    Mean & 219 & 17 & 37 & 10 & 2 \\
    Std & 133 & 26 & 11 & 12 & 5 \\
    Min & 32 & 2 & 7 & 1 & 0 \\
    25\% & 99 & 6 & 29 & 3 & 0 \\
    50\% & 204 & 9 & 38 & 7 & 1 \\
    75\% & 304 & 16 & 46 & 11 & 2 \\
    Max & 500 & 168 & 50 & 63 & 22
  \\ \hline \\
  \end{tabular}
  \label{table:characteristics}
\end{table}

\section{PMLBmini Exemplary Usage} \label{sec:appendix_example}

Below, we illustrate how researchers and practitioners can use our benchmarking tool for empirical comparisons and meta-feature analyses with their own tabular classifier in the low-data regime:

\begin{lstlisting}[language=python]
from yourpackage import YourEstimator
import tabmini

# Load the dataset
# Tabmini also provides a dummy dataset for testing purposes, you can load it with tabmini.load_dummy_dataset() 
# If reduced is set to True, the dataset will exclude all the data that has been used to develop TabPFN's prior
dataset = tabmini.load_dataset(reduced=False)

# Prepare the estimator you want to benchmark against the other estimators
estimator = YourEstimator()

# Perform the comparison
train_results, test_results = tabmini.compare(
    "MyEstimator",
    estimator,
    dataset,
    working_directory=pathlib.Path.cwd() / "results",
    scoring_method="roc_auc",
    cv=3,
    time_limit=3600,
    device="cpu"
)

# Generate the meta-feature analysis
meta_features = tabmini.get_meta_feature_analysis(dataset, test_results, "MyEstimator", correlation_method="spearman")

# Save the results and meta-feature analysis to a CSV file
test_results.to_csv("results.csv")
meta_features.to_csv("meta_features.csv")
\end{lstlisting}

\section{Additional Results} \label{sec:appendix_results}

\begin{figure}[h]
\centering
\includegraphics[width=\textwidth]{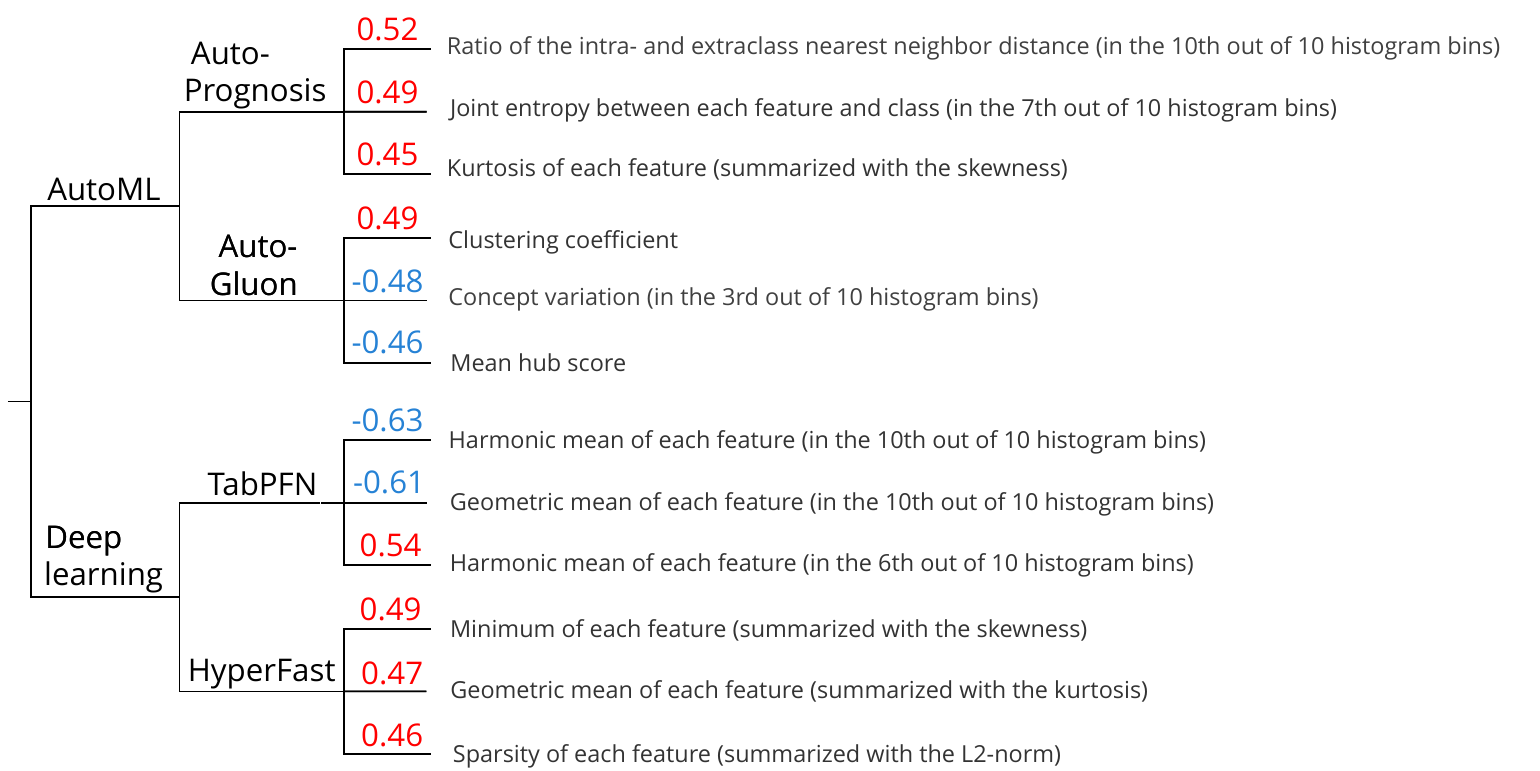}
\caption{Top-3 meta-features per approach. We computed all PyMFE meta-features per dataset, the mean test AUC differences between each AutoML / deep learning method and logistic regression per dataset, the absolute Spearman rank correlation coefficient between each PyMFE meta-feature and the performance difference across datasets (Sect.~\ref{sec:interface}); and finally selected the top-3 meta-features with the largest absolute correlations. Positive correlation coefficients are shown in red, negative correlation coefficients in blue.}
\label{fig:results_d}
\end{figure}

In this section, we present additional experimental results in Table~\ref{table:results} and~\ref{table:results_gbdt} and show the top-3 meta-features for each AutoML and deep learning method in Fig.~\ref{fig:results_d}.

\tabcolsep=0.13cm
\begin{longtable}{p{2.3cm} p{0.5cm} p{0.5cm} p{1.6cm} p{1.6cm} p{1.6cm} p{1.6cm} p{1.6cm} p{0.6cm}}
\caption{Training AUC (mean test AUC) and $\lambda^*$ across all 44 datasets in PMLBmini with sample size M and feature set size N, ordered for sample size ranges in steps of 100.}\\
\label{table:results}
\multirow{2}{=}[1em]{\bf PMLBmini dataset} & \multirow[t]{2}{=}{\bf M} & \multirow[t]{2}{=}{\bf N} & \multicolumn{2}{c}{\bf AutoML} & \multicolumn{2}{c}{\bf Deep learning} & \multirow{2}{=}{\bf Logistic regression} & \multirow[t]{2}{=}{$\boldsymbol{\lambda^*}$} \\
& & & \textbf{Auto-Prognosis} & \textbf{Auto-Gluon} & \textbf{TabPFN} & \textbf{HyperFast} & & 
\\ \hline \\
parity5 & 32 & 5 & 0.50 (0.27) & 0.04 (\textbf{0.98}) & 1.00 (0.02) & 1.00 (0.02) & 0.50 (0.17) & 0.5\\
analcatdata\_ fraud & 42 & 11 & 0.93 (\textbf{0.86}) & 0.99 (0.68) & 1.00 (0.79) & 0.99 (0.73) & 0.89 (0.77) & 0.5\\
analcatdata\_ aids & 50 & 4 & 1.00 (\textbf{0.73}) & 0.94 (0.67) & 1.00 (0.63) & 0.80 (0.53) & 0.78 (0.61) & 0.004\\
analcatdata\_ bankruptcy & 50 & 6 & 1.00 (\textbf{0.98}) & 1.00 (0.97) & 1.00 (0.96) & 0.99 (0.88) & 0.99 (0.97) & 0.004\\
analcatdata\_ japansolvent & 52 & 9 & 1.00 (0.85) & 0.99 (0.88) & 1.00 (\textbf{0.91}) & 0.97 (\textbf{0.91}) & 0.94 (0.85) & 0.1\\
labor & 57 & 16 & 1.00 (0.88) & 1.00 (0.95) & 1.00 (\textbf{0.99}) & 1.00 (0.98) & 1.00 (0.97) & 0.02\\
analcatdata\_ asbestos & 83 & 3 & 0.87 (\textbf{0.87}) & 0.89 (0.85) & 0.93 (0.85) & 0.87 (\textbf{0.87}) & 0.87 (0.86) & 0.5\\
lupus & 87 & 3 & 0.92 (0.84) & 0.86 (0.77) & 0.86 (0.82) & 0.83 (0.79) & 0.85 (\textbf{0.85}) & 0.1 \\
postoperative\_ patient\_data & 88 & 8 & 0.59 (\textbf{0.49}) & 0.12 (0.46) & 0.99 (0.44) & 0.87 (0.34) & 0.65 (0.38) & 0.5\\
analcatdata\_ cyyoung9302 & 92 & 10 & 1.00 (\textbf{0.89}) & 0.99 (0.85) & 0.99 (0.87) & 0.96 (0.84) & 0.94 (0.87) & 0.1\\
analcatdata\_ cyyoung8092 & 97 & 10 & 0.91 (0.73) & 0.99 (\textbf{0.87}) & 0.98 (0.85) & 0.91 (0.84) & 0.93 (0.79) & 0.1\\
analcatdata\_ creditscore & 100 & 6 & 1.00 (\textbf{1.00}) & 1.00 (0.99) & 1.00 (\textbf{1.00}) & 0.94 (0.87) & 0.97 (0.94) & 0.02\\
\\ \hline \\
\multicolumn{3}{l}{median M = 32, ..., 100} & 0.97 (\textbf{0.86}) & 0.99 (\textbf{0.86}) & 1.00 (0.85) & 0.95 (0.84) & 0.91 (0.85) & \\
\\ \hline \\
appendicitis & 106 & 7 & 0.88 (0.78) & 0.91 (0.85) & 0.97 (0.82) & 0.86 (\textbf{0.87}) & 0.86 (0.84) & 0.5\\
molecular\_bio-logy\_promoters & 106 & 57 & 1.00 (0.88) & 1.00 (\textbf{0.91}) & 1.00 (0.88) & 1.00 (0.89) & 1.00 (0.88) & 0.5\\
analcatdata\_ boxing1 & 120 & 3 & 0.97 (\textbf{0.89}) & 0.96 (0.85) & 0.99 (0.76) & 0.72 (0.67) & 0.68 (0.67) & 0.5\\
mux6 & 128 & 6 & 0.50 (\textbf{1.00}) & 1.00 (\textbf{1.00}) & 1.00 (\textbf{1.00}) & 1.00 (0.95) & 0.78 (0.70) & 0.5\\
analcatdata\_ boxing2 & 132 & 3 & 0.92 (\textbf{0.82}) & 0.91 (0.75) & 0.85 (0.71) & 0.75 (0.70) & 0.70 (0.68) & 0.5\\
hepatitis & 155 & 19 & 0.87 (\textbf{0.85}) & 0.99 (0.80) & 0.99 (\textbf{0.85}) & 0.92 (0.83) & 0.93 (0.84) & 0.5\\
corral & 160 & 6 & 1.00 (\textbf{1.00}) & 1.00 (\textbf{1.00}) & 1.00 (\textbf{1.00}) & 1.00 (\textbf{1.00}) & 0.97 (0.96) & 0.1\\
glass2 & 163 & 9 & 1.00 (0.89) & 1.00 (\textbf{0.91}) & 1.00 (0.89) & 0.89 (0.79) & 0.81 (0.72) & 0.004\\
backache & 180 & 32 & 0.90 (0.60) & 1.00 (0.71) & 1.00 (0.75) & 0.93 (\textbf{0.78}) & 0.90 (0.72) & 0.5\\
prnn\_crabs & 200 & 7 & 1.00 (\textbf{1.00}) & 1.00 (\textbf{1.00}) & 1.00 (\textbf{1.00}) & 0.83 (0.81) & 1.00 (\textbf{1.00}) & 0.004\\
\\ \hline \\
\multicolumn{3}{l}{median M = 101, ..., 200} & 0.95 (\textbf{0.89}) & 1.00 (0.88) & 1.00 (0.87) & 0.91 (0.82) & 0.88 (0.78)\\
\\ \hline \\
sonar & 208 & 60 & 1.00 (0.88) & 1.00 (\textbf{0.92}) & 1.00 (\textbf{0.92}) & 0.95 (0.89) & 0.95 (0.85) & 0.5\\
biomed & 209 & 8 & 1.00 (\textbf{1.00}) & 1.00 (0.96) & 1.00 (0.95) & 0.96 (0.93) & 0.96 (0.94) & 0.004\\
prnn\_synth & 250 & 2 & 0.98 (0.94) & 0.96 (\textbf{0.95}) & 0.97 (\textbf{0.95}) & 0.93 (0.94) & 0.94 (0.94) & 0.02\\
analcatdata\_ lawsuit & 264 & 4 & 1.00 (0.99) & 1.00 (0.99) & 1.00 (\textbf{1.00}) & 0.99 (0.98) & 1.00 (\textbf{1.00}) & 0.004\\
spect & 267 & 22 & 0.87 (\textbf{0.84}) & 0.94 (0.81) & 0.95 (0.83) & 0.90 (0.83) & 0.90 (0.82) & 0.5\\
heart\_statlog & 270 & 13 & 0.94 (\textbf{0.91}) & 0.97 (0.87) & 0.98 (0.90) & 0.93 (0.89) & 0.93 (0.89) & 0.5\\
breast\_cancer & 286 & 9 & 0.76 (0.69) & 0.96 (0.67) & 0.90 (\textbf{0.73}) & 0.80 (0.69) & 0.73 (0.70) & 0.5\\
heart\_h & 294 & 13 & 0.92 (0.87) & 0.96 (0.87) & 0.97 (\textbf{0.88}) & 0.89 (0.85) & 0.88 (0.86) & 0.5\\
hungarian & 294 & 13 & 0.99 (\textbf{0.86}) & 1.00 (0.85) & 0.96 (\textbf{0.86}) & 0.92 (0.84) & 0.89 (0.85) & 0.5\\
\\ \hline \\
\multicolumn{3}{l}{median M = 201, ..., 300} & 0.98 (0.88) & 0.97 (0.87) & 0.97 (\textbf{0.90}) & 0.93 (0.89) & 0.93 (0.86)\\
\\ \hline \\
cleve & 303 & 13 & 0.96 (\textbf{0.90}) & 0.98 (\textbf{0.90}) & 0.99 (0.89) & 0.93 (0.88) & 0.90 (0.88) & 0.5\\
heart\_c & 303 & 13 & 0.94 (\textbf{0.91}) & 0.95 (0.90) & 0.98 (\textbf{0.91}) & 0.94 (0.89) & 0.92 (\textbf{0.91}) & 0.5\\
haberman & 306 & 3 & 0.86 (0.70) & 0.76 (0.71) & 0.82 (\textbf{0.72}) & 0.65 (0.58) & 0.70 (0.66) & 0.5\\
bupa & 345 & 5 & 0.70 (0.66) & 0.77 (0.65) & 0.72 (\textbf{0.68}) & 0.72 (0.66) & 0.68 (0.67) & 0.1\\
spectf & 349 & 44 & 1.00 (0.91) & 1.00 (\textbf{0.94}) & 1.00 (0.93) & 0.91 (0.87) & 0.94 (0.88) & 0.1\\
ionosphere & 351 & 34 & 1.00 (0.97) & 1.00 (\textbf{0.98}) & 1.00 (\textbf{0.98}) & 0.96 (0.97) & 0.97 (0.90) & 0.5\\
colic & 368 & 22 & 0.99 (\textbf{0.87}) & 0.98 (\textbf{0.87}) & 1.00 (\textbf{0.87}) & 0.90 (0.86) & 0.89 (0.86) & 0.5\\
horse\_colic & 368 & 22 & 0.99 (\textbf{0.88}) & 0.98 (0.85) & 1.00 (0.84) & 0.89 (0.83) & 0.87 (0.82) & 0.5\\
\\ \hline \\
\multicolumn{3}{l}{median M = 301, ..., 400} & 0.98 (\textbf{0.89}) & 0.98 (\textbf{0.89}) & 1.00 (0.88) & 0.91 (0.87) & 0.90 (0.87)\\
\\ \hline \\
house\_votes\_84 & 435 & 16 & 1.00 (\textbf{0.99}) & 1.00 (\textbf{0.99}) & 1.00 (\textbf{0.99}) & 0.99 (0.98) & 0.99 (\textbf{0.99}) & 0.1\\
vote & 435 & 16 & 1.00 (\textbf{1.00}) & 1.00 (0.99) & 1.00 (\textbf{1.00}) & 0.99 (0.99) & 1.00 (0.99) & 0.5\\
saheart & 462 & 9 & 0.82 (\textbf{0.77}) & 0.69 (0.75) & 0.83 (\textbf{0.77}) & 0.81 (0.76) & 0.79 (\textbf{0.77}) & 0.5\\
clean1 & 476 & 168 & 1.00 (0.93) & 1.00 (\textbf{1.00}) & 1.00 (0.99) & 0.98 (0.96) & 1.00 (\textbf{1.00}) & 0.004\\
irish & 500 & 5 & 1.00 (\textbf{1.00}) & 1.00 (\textbf{1.00}) & 1.00 (\textbf{1.00}) & 0.98 (0.97) & 0.85 (0.83) & 0.1\\
\\ \hline \\
\multicolumn{3}{l}{median M = 401, ..., 500} & 1.00 (\textbf{0.99}) & 1.00 (\textbf{0.99}) & 1.00 (\textbf{0.99}) & 0.98 (0.97) & 0.99 (\textbf{0.99})\\
\\ \hline \\
\end{longtable}

\begin{table}[ht]
\centering
\caption{Mean test AUC for gradient-boosted decision trees across all 44 datasets in PMLBmini, ordered for sample size. LightGBM reaches a similar median performance to logistic regression; the median performance for XGBoost and CatBoost is worse than for logistic regression, AutoML, and deep neural networks, though.}
\label{tab:tree-models}
\begin{tabular}{lrrr}
\textbf{PMLBmini dataset} & \textbf{LightGBM} & \textbf{XGBoost} & \textbf{CatBoost} \\
\midrule
parity5 & \textbf{0.50} & 0.19 & 0.21 \\
analcatdata\_fraud & 0.50 & \textbf{0.66} & 0.53 \\
analcatdata\_aids & 0.50 & \textbf{0.73} & 0.59 \\
analcatdata\_bankruptcy & 0.50 & \textbf{0.86} & 0.84 \\
analcatdata\_japansolvent & 0.50 & 0.73 & \textbf{0.80} \\
labor & 0.50 & \textbf{0.74} & 0.69 \\
analcatdata\_asbestos & \textbf{0.82} & 0.77 & 0.80 \\
lupus & 0.73 & 0.74 & \textbf{0.77} \\
postoperative\_patient\_data & 0.36 & 0.47 & \textbf{0.50} \\
analcatdata\_cyyoung9302 & \textbf{0.84} & 0.71 & 0.77 \\
analcatdata\_cyyoung8092 & \textbf{0.78} & 0.74 & 0.68 \\
analcatdata\_creditscore & 0.94 & \textbf{0.99} & 0.97 \\
appendicitis & \textbf{0.78} & 0.77 & 0.75 \\
molecular\_biology\_promoters & \textbf{0.92} & 0.80 & 0.70 \\
analcatdata\_boxing1 & \textbf{0.69} & \textbf{0.69} & 0.61 \\
mux6 & \textbf{0.96} & 0.56 & 0.67 \\
analcatdata\_boxing2 & \textbf{0.78} & \textbf{0.78} & 0.76 \\
hepatitis & \textbf{0.78} & 0.69 & 0.62 \\
corral & \textbf{1.00} & 0.92 & 0.90 \\
glass2 & \textbf{0.92} & 0.75 & 0.70 \\
backache & \textbf{0.66} & 0.50 & 0.53 \\
prnn\_crabs & \textbf{0.97} & 0.81 & 0.78 \\
sonar & \textbf{0.91} & 0.71 & 0.71 \\
biomed & \textbf{0.95} & 0.85 & 0.73 \\
prnn\_synth & \textbf{0.94} & 0.86 & 0.86 \\
analcatdata\_lawsuit & \textbf{0.99} & 0.91 & 0.63 \\
spect & \textbf{0.80} & 0.61 & 0.50 \\
heart\_statlog & \textbf{0.86} & 0.74 & 0.78 \\
breast\_cancer & \textbf{0.66} & 0.59 & 0.55 \\
heart\_h & \textbf{0.86} & 0.78 & 0.77 \\
hungarian & \textbf{0.84} & 0.81 & 0.77 \\
cleve & \textbf{0.86} & 0.75 & 0.75 \\
heart\_c & \textbf{0.88} & 0.79 & 0.76 \\
haberman & \textbf{0.70} & 0.59 & 0.57 \\
bupa & \textbf{0.63} & 0.59 & 0.59 \\
spectf & \textbf{0.91} & 0.69 & 0.73 \\
ionosphere & \textbf{0.97} & 0.85 & 0.83 \\
colic & \textbf{0.85} & 0.83 & 0.81 \\
horse\_colic & \textbf{0.87} & 0.83 & 0.80 \\
house\_votes\_84 & \textbf{0.99} & 0.96 & 0.96 \\
vote & \textbf{0.99} & 0.96 & 0.95 \\
saheart & \textbf{0.70} & 0.65 & 0.67 \\
clean1 & \textbf{1.00} & \textbf{1.00} & \textbf{1.00} \\
irish & \textbf{1.00} & \textbf{1.00} & \textbf{1.00} \\
\bottomrule
\end{tabular}
\label{table:results_gbdt}
\end{table}

\end{document}